\documentclass[10pt,journal,compsoc]{IEEEtran}

\ifCLASSOPTIONcompsoc
  \usepackage[nocompress]{cite}
\else
  \usepackage{cite}
\fi

\ifCLASSINFOpdf
\else
\fi

\usepackage{graphicx}
\usepackage[parfill]{parskip}
\usepackage{amsmath,amssymb,amstext}
\usepackage{mathdots}
\usepackage{algorithm}
\usepackage[noend]{algpseudocode}
\usepackage{multirow}
\usepackage{caption}

\begin{document}
\title{Affine and Regional Dynamic Time Warping}
\author{Tsu-Wei~Chen,~Meena~Abdelmaseeh,~and~Daniel~Stashuk

\IEEEcompsocitemizethanks{\IEEEcompsocthanksitem T. Chen, M. Abdelmaseeh and D. Stashuk are with the Department
of Systems Design Engineering, University of Waterloo, Waterloo,
Canada.}
}

\IEEEtitleabstractindextext{%
\begin{abstract}
Pointwise matches between two time series are of great importance in time series analysis, and dynamic time warping (DTW) is known to provide generally reasonable matches. There are situations where time series alignment should be invariant to scaling and offset in amplitude or where local regions of the considered time series should be strongly reflected in pointwise matches. Two different variants of DTW, affine DTW (ADTW) and regional DTW (RDTW), are proposed to handle scaling and offset in amplitude and provide regional emphasis respectively. Furthermore, ADTW and RDTW can be combined in two different ways to generate alignments that incorporate advantages from both methods, where the affine model can be applied either globally to the entire time series or locally to each region. The proposed alignment methods outperform DTW on specific simulated datasets, and one-nearest-neighbor classifiers using their associated difference measures are competitive with the difference measures associated with state-of-the-art alignment methods on real datasets.
\end{abstract}

}

\maketitle

\IEEEdisplaynontitleabstractindextext
\IEEEpeerreviewmaketitle

\section{Introduction}
\IEEEPARstart{A}{} time series is a sequence of values that are typically arranged in a chronological order, and data of such form is abundant in everyday life. Discovery of a set of matches between points in two time series can be tremendously useful for analysis. If point $a$ in time series $s$ is of high interest, researchers may also be interested in finding the point that best matches point $a$ in another time series $t$. Points of interest include hints of financial meltdown from stock market prices, regulatory genes from gene expression data and earthquake activities from seismic data.

\begin{figure}[!h]
\begin{center}
\includegraphics[scale = 0.5]{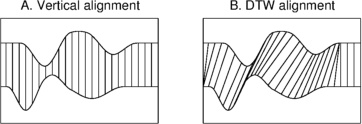}
\caption{Vertical vs DTW alignment.}
\label{Figure:DTWConcept}
\end{center}
\end{figure}

Dynamic time warping (DTW) is a method that matches points in two time series based on the assumption that non-linear temporal variations exist. Figure \ref{Figure:DTWConcept} illustrates a purely vertical alignment (an alignment is defined to be a set of matches) and the DTW alignment of two time series subject to non-linear temporal variations, where a match between two points from different time series is illustrated by connecting the two points with a line. The DTW alignment is much more visually consistent. This work focuses on DTW, because it is a widely known method for aligning two time series that has enjoyed success in many domains. In addition, a comprehensive survey demonstrated that DTW outperforms many other methods across many applications \cite{ding2008querying}. However, DTW can produce pathological alignments, which are identified based on a context wherein the identifier has a particular model in mind. As a result, there has been a large influx of methods based on DTW to realize particular models for obtaining better alignments under specific contexts \cite{fu2011review}.

\begin{figure}[!h]
\begin{center}
\includegraphics[scale = 0.37]{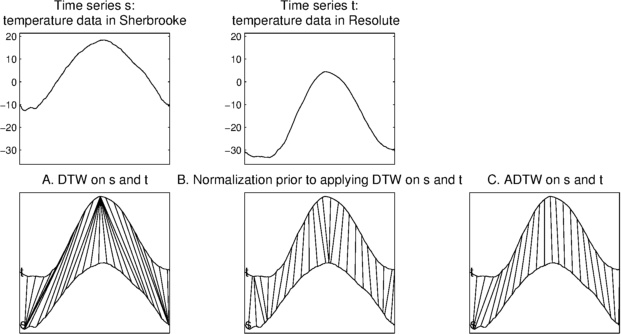}
\caption{DTW vs ADTW alignment of daily temperature across a year in Sherbrooke and Resolute, Canada.}
\label{Figure:ADTWApplyOnIntroductoryExample}
\end{center}
\end{figure}

In this work, two alignment methods that add specific models to DTW are proposed: affine DTW (ADTW) and regional DTW (RDTW). ADTW models one time series as an amplitude-scaled and offset-biased version of another time series during alignment. It tries to find the best scaling, offset and alignment simultaneously. Numerous types of time series fall under this affine model. For example, temperature and humidity data can be subject to different scalings, offsets and temporal variations depending on the geographic location and environment. An unintuitive alignment produced by DTW for two temperature time series subject to scaling, offset and temporal variation is illustrated in Figure  \ref{Figure:ADTWApplyOnIntroductoryExample}A, where DTW matches a large number of points in one time series to the peak in another time series. While normalizing $s$ and $t$ before applying DTW can alleviate this undesired behavior to some extent, ADTW nonetheless provides the most visually consistent alignment (see Figure \ref{Figure:ADTWApplyOnIntroductoryExample}B and C). ADTW is a simplified version of the method proposed in \cite{qiao2006affine}, where scaling, offset, rotation, and shear/squeeze mappings are imposed on images when applying DTW. ADTW only models scaling and offset, so its alignment will not be confused by modeling of rotation and shear/squeeze mappings for time series that cannot undergo such transformations.

\begin{figure}[!h]
\begin{center}
\includegraphics[scale = 0.35]{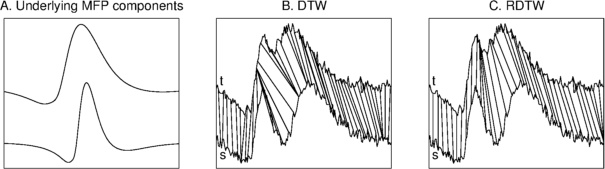}
\caption{DTW vs RDTW alignment of two MUPs with different degrees of MFP overlap.}
\label{Figure:RDTWApplyOnIntroductoryExample}
\end{center}
\end{figure}

Scenarios may arise where there are local regions in a time series reflective of components of interest, so they should be emphasized to find a more desirable alignment between two time series. Regional DTW (RDTW) is proposed to accommodate this scenario by substituting the pointwise distance in DTW with a regional distance. Many types of time series contain components of interest that should be focused on. For example, a motor unit potential (MUP) is the ensemble summation of several muscle fiber potentials (MFPs), and their analysis is crucial to determining the characteristics of the MUP. In Figure \ref{Figure:RDTWApplyOnIntroductoryExample}, each MUP (time series $s$ and $t$ respectively) is the ensemble summation of two MFPs. Each MFP can be shifted in time by a different amount and can thus be subject to different degrees of overlap. DTW produces a bad alignment where a large portion of the leftmost MFP in $t$ is matched to the rightmost MFP in $s$ (see Figure \ref{Figure:RDTWApplyOnIntroductoryExample}B). In contrast, RDTW aligns the constituent MFP contributions in a more desirable manner (see Figure \ref{Figure:RDTWApplyOnIntroductoryExample}C).

ADTW and RDTW can be combined to include both affine modeling and emphasis on local regions. Two different ways of combining ADTW and RDTW (one in a global manner and one in a local manner) are proposed. Global-affine RDTW (GARDTW) models one time series as a scaled and offset version of another time series when aligning them with regional emphasis. For example, the amount of rainfall over time is subject to different scalings, offsets and temporal variations across different locations. In the analysis of this rainfall data, placing an emphasis on sections with short but heavy amounts of rainfall can be useful in predicting such behaviors. The preference for this example is to emphasize on matching components reflective of short but heavy amounts of rainfall within two time series that can undergo scaling, offset and temporal variation. Local-affine RDTW (LARDTW) emphasizes on regions when performing an alignment, where each region in one time series is modeled as a scaled and offset version of the respective matched region in another time series. Revisiting the MUP example, the same MFP can have different scalings caused by slight electrode movement. The objective in this example is to correctly align the MFPs that can undergo different scalings within two MUPs.

The rest of this paper is organized as follows. Section 2 provides a review for the technical details of DTW, and the proposed methods of ADTW, RDTW, GARDTW and LARDTW are described in detail. Section 3 covers evaluations of the alignments and difference measures generated by the proposed methods. Finally, Section 4 concludes this paper. All figures and results in this paper can be easily reproduced using the publicly available code at \cite{SupportingWebpage}.

\section{Alignment Methodology}

\subsection{Notation}
Let $s=(s_1,s_2,...,s_n)\in \mathbb{R}^{n}$ and $t=(t_1,t_2,...,t_m)\in \mathbb{R}^{m}$ be two time series of interest. Also, let $p$ represent a sequence of matched points between $s$ and $t$, where 
\begin{align*}
p=\{p(1)&=(a_1,b_1),p(2)=(a_2,b_2),...,\\
&\qquad \qquad \qquad p(|p|)=(a_{|p|},b_{|p|})\}
\end{align*}
 and $(a_k,b_k)\in \mathbb{Z}_{>0}^2$ means that point $s_{a_k}$ is matched to point $t_{b_k}$. In addition, let $d$ be a difference measure between two points, and $d$ is assumed to be the squared difference unless mentioned otherwise.

\subsection{DTW Review}
DTW is a method that matches points in two time series that are subject to non-linear temporal variations.  For a pair of time series $s$ and $t$, DTW searches for an optimal alignment $p^*$ among all possible alignments $p\in \mathbb{P}$ such that $$D(s,t,p)=\sum_{k=1}^{|p|}d(s_{a_k},t_{b_k})$$ is minimized subject to the following constraints:
\begin{itemize}
\item Boundary: $p(1)=(1,1)$ and $p(|p|)=(n,m)$.
\item Monotonicity: If $p(k)=(a,b)$ and $p(k+1)=(c,d)$, then $c\geq a $ and $d\geq b$ $\forall k$.
\item Step Size: If $p(k)=(a,b)$ and $p(k+1)=(c,d)$, then $c-a\leq 1$ and $d-b\leq 1$ $\forall k$.
\end{itemize}
For simplicity, the boundary, monotonicity and step size constraints will be jointly referred to as the DTW constraints. $D(s,t,p^*)$ will also be referred to as the DTW difference measure.

Dynamic programming is effective in reducing the time complexity for finding the optimal alignment $p^*$ to this constrained optimization problem, because $p^*$ has optimal substructures
and there are overlapping subproblems. A solution can be formulated using these properties. Let $p_{(a,b)}^*$ be the optimal alignment for $(s_1,s_2,...,s_a)$ and $(t_1,t_2,...,t_b)$ subject to the DTW constraints. First, a table of $D(s,t,p_{(i,j)}^*)$ values is constructed for all $1\leq i\leq n$ and $1\leq j\leq m$ where $i$ is the row position and $j$ is the column position. This table is referred to as the DTW table, and $p^*$ can be found after building this table. The DTW table can be updated starting from the first row ($i=1$) from left to right ($j=1$ to $j=m$), and then the next row ($i=2$) can be filled from left to right as well until the $n^\text{th}$ row is reached. The update formula is as follows:
\begin{align}
\label{Equation:DTWUpdate}
D(s,t,p_{(a,b)}^*) &= d(s_a,t_b) + \text{min}(D(s,t,p_{(a-1,b-1)}^*), \nonumber \\ 
& \qquad \qquad D(s,t,p_{(a,b-1)}^*), D(s,t,p_{(a-1,b)}^*))
\end{align}

After constructing the DTW table, a backtracking procedure can be applied on the DTW table starting from $D(s,t,p_{(n,m)}^*)$ to generate the optimal alignment $p^*$.

Additional constraints can be placed on alignments to 1) eliminate pathological alignments, and 2) reduce time and space complexity. A well-known constraint is the Sakoe-Chiba band, where $s_i$ can only be matched to points from the set $\{t_{i-w_q},t_{i-w_q+1},...,t_i,...,t_{i+w_q-1},t_{i+w_q}\}$ and $w_q\in \mathbb{Z}_{\geq 0}$. The width of the band $w_b$ is given by $w_b=1+2w_q$, and it is recommended to be tuned based on specific problems \cite{ratanamahatana2004everything}.

For simplicity of complexity analysis, let us assume that $n=m$. Since the update formula in Equation \ref{Equation:DTWUpdate} for filling each element in the DTW table is $O(1)$ in time, filling the entire table requires a time complexity of $O(w_bn)$ under the Sakoe-Chiba band. The space complexity for constructing the DTW table is $O(w_bn)$ as well. The longest path that can be obtained from backtracking is $2n-1$, so backtracking is $O(n)$ in time and space. Hence, the total time and space complexities for finding the optimal alignment $p^*$ are $O(w_bn)$.

\subsection{Affine DTW}
Affine DTW (ADTW) increments DTW to allow arbitrary scaling and offset in amplitude between two time series subject to temporal variations. In ADTW, $s$ is assumed to be a scaled and offset version of $t$ with temporal variations. In more formal terms, the goal is to find a path $p^*$, scaling $c^*\in \mathbb{R}$ and offset $e^*\in \mathbb{R}$ that minimize
$$D_A(s,t,p,c,e)=\sum_{k=1}^{|p|}d(s_{a_k},ct_{b_k}+e)$$
subject to the DTW constraints. For brevity, $D_A(s,t,p,c,e)$ subject to the DTW constraints will be referred to as $D_A(s,t,p,c,e)_\text{constr.}$. This formulation aims to optimize for a global minimum in $D_A(s,t,p,c,e)_\text{constr.}$ with respect to $p$, $c$ and $e$ simultaneously, which is different from finding the scaling and offset first prior to applying DTW to obtain an alignment.

Finding $(p^*,c^*,e^*)$ is too computationally expensive because dynamic programming can no longer be applied, so hard expectation-maximization (EM) is used to find a suboptimal solution $(p^l,c^l,e^l)$ in Algorithm \ref{Algorithm:ADTW}. Hard EM guarantees that $D_A(s,t,p_{v+1}^l,c_{v+1}^l,e_{v+1}^l)_{\text{constr.}}\leq D_A(s,t,p_v^l,c_v^l,e_v^l)_{\text{constr.}}$, and it converges to a local optimum at a linear rate when certain conditions are fulfilled \cite{celeux1992classification}.
\begin{algorithm}
\caption{ADTW}\label{Algorithm:ADTW}
\begin{algorithmic}[1]
\State $p^l, c^l, e^l, c_0^l \gets 1, e_0^l \gets 0, D_{A,\text{prev}} \gets \infty, v \gets 1$
\While{$1$}
\State $p_{v}^l \gets \underset{p}{\text{argmin}} \ D_A(s,t,p,c_{v-1}^l,e_{v-1}^l)_\text{constr.}$
\State $(c_v^l,e_v^l) \gets \underset{c,e}{\text{argmin}} \ D_A(s,t,p_v^l,c,e)_\text{constr.}$
\If {$D_{A,\text{prev}}-D_A(s,t,p^l,c^l,e^l) < D_\text{stop}$}
\State $p^l \gets p_v^l, c^l \gets c_v^l, e^l \gets e_v^l$
\State \textbf{break}
\EndIf
\State $v \gets v + 1$
\EndWhile
\end{algorithmic}
\end{algorithm}

In Algorithm \ref{Algorithm:ADTW}, $p_v^l$ is obtained by applying DTW on $s$ and $c_{v-1}^lt+e_{v-1}^l$, and $(c_v^l,e_v^l)$ are computed with the following equations after setting $p=p_v^l$:
\begin{equation}
\label{Equation:ADTWScaling}
c_v^l=\frac{\sum_{k=1}^{|p|}s_{a_k}t_{b_k}-\frac{1}{|p|}(\sum_{k=1}^{|p|}s_{a_k})(\sum_{k=1}^{|p|}t_{b_k})}{\sum_{k=1}^{|p|}t_{b_k}^2-\frac{1}{|p|}(\sum_{k=1}^{|p|}t_{b_k})^2}
\end{equation}
\begin{equation}
\label{Equation:ADTWOffset}
e_v^l=\frac{1}{|p|}\sum_{k=1}^{|p|}(s_{a_k}-c_v^lt_{b_k})
\end{equation}
The above equations can be derived in a manner similar to linear least squares. The scaling $c_v^l$ can be constrained to exist between $c_\text{min}$ and $c_\text{max}$ to avoid improbable scalings. In addition, $D_A(s,t,p^l,c^l,e^l)$ will be referred to as the ADTW difference measure. Note that a more general version of ADTW where each prespecified subset of points has its own scaling and offset can be solved in a similar way.

Assuming $n=m$, each iteration in Algorithm \ref{Algorithm:ADTW} takes $O(w_bn)$ time and space to run DTW to obtain $p_v^l$. Looking at Equation \ref{Equation:ADTWScaling} and \ref{Equation:ADTWOffset}, computing $(c_v^l,e_v^l)$ is $O(n)$ in time and $O(1)$ in space. Hence, ADTW is $O(n_cw_bn)$ in time and $O(w_bn)$ in space, where $n_c$ is the number of iterations for convergence.

\subsection{Regional DTW}
Regional DTW (RDTW) modifies DTW to place more weight in a region of points potentially representative of a component of interest in a time series. This is accomplished by substituting the pointwise distance measure $d$ with a distance $d_r$ that measures the difference between points in a region. Let $w_r=1+2w_h\in \mathbb{Z}_{\geq 1}$ be the region width to consider. Then, RDTW finds an alignment $p^*$ that minimizes
$$D_R(s,t,p,w_h)=\sum_{k=1}^{|p|}d_r(s_{a_k},t_{b_k},w_h)$$
subject to the DTW constraints, where
$$d_r(s_a,t_b,w_h)=\frac{1}{w_{a,b}}\sum_{\substack{w=-w_h\\1\leq a+w\leq n\\1\leq b+w\leq m}}^{w_h}d(s_{a+w},t_{b+w})$$
and $w_{a,b}$ is the number of distances added in the above summation.
Dynamic programming can be utilized in the same manner as DTW, where the update formula is as follows:
\begin{align}
\label{Equation:RDTWUpdate}
& D_R(s,t,p_{(a,b)}^*,w_h) \nonumber \\
&= d_r(s_a,t_b,w_h) + \text{min}(D_R(s,t,p_{(a-1,b-1)}^*,w_h), \nonumber \\
 & \qquad D_R(s,t,p_{(a,b-1)}^*,w_h), D_R(s,t,p_{(a-1,b)}^*,w_h))
\end{align}
The same DTW techniques can be used to construct the RDTW table and obtain the optimal alignment $p^*$. $D_R(s,t,p^*,w_h)$ is referred to as the RDTW difference measure.

Assuming $n=m$, the RDTW table requires $O(w_bn)$ elements to be filled with the update formula. It turns out that most elements in the RDTW table can be updated with a time complexity of $O(1)$ instead of $O(w_r)$ using the following observation:
\begin{align*}
\label{Equation:RDTWElementConstantTimeUpdate}
& d_r(s_{a},t_{b},w_h) = \frac{1}{w_{a,b}}[-d(s_{a-w_h-1},t_{b-w_h-1}) + \\ & \qquad \qquad w_{a-1,b-1}d_r(s_{a-1},t_{b-1},w_h) + d(s_{a+w_h},t_{b+w_h})]
\end{align*}

This observation is only applicable when $s_{a-1}$ and $t_{b-1}$ exist, which corresponds to $w_b(n-1)$ elements. The remaining $w_b$ elements take $O(w_r)$ time, so the total time complexity is $O(w_b(n-1))+O(w_bw_r)=O(w_bn)$ because $w_r\leq n$. The total space complexity is also $O(w_bn)$.

\begin{figure}[!h]
\begin{center}
\includegraphics[scale = 0.38]{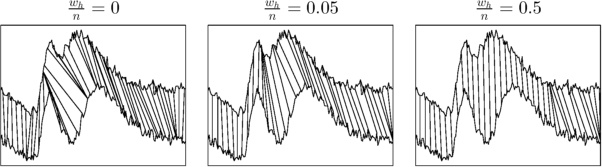}
\caption{DTW alignments with different region widths.}
\label{Figure:RDTWEffectsOfWidth}
\end{center}
\end{figure}

So far, the effects of the region width $w_r$ have not been discussed, which is crucial to achieving good results. RDTW with different $w_r$'s is applied to the same MUP alignment example from the introduction in Figure \ref{Figure:RDTWEffectsOfWidth}. Highly variable alignments are observed across the different widths, and the alignment is most reasonable when $\frac{w_h}{n}=0.05$. Setting $w_h$ by searching for the value that offers the best result based on a target evaluation criterion is proposed.

\subsection{Global-Affine RDTW}
In global-affine RDTW (GARDTW), one time series is modeled as the scaled and offset version of another time series when aligning them with regional emphasis. Formally, the goal is to find a path $p^*$, scaling $c^*$ and offset $e^*$ that minimize
$$D_G(s,t,p,c,e,w_h)=\sum_{k=1}^{|p|}d_g(s_{a_k},t_{b_k},c,e,w_h)$$
subject to the DTW constraints, and
\begin{align*}
&d_g(s_{a_k},t_{b_k},c,e,w_h) \\ 
&=\frac{1}{w_{a_k,b_k}}\sum_{\substack{w=-w_h\\w:1\leq a_k+w\leq n\\w:1\leq b_k+w\leq m}}^{w_h}d(s_{a_k+w},ct_{b_k+w}+e)
\end{align*}
$D_G(s,t,p,c,e,w_h)$ subject to the DTW constraints will be referred to as $D_G(s,t,p,c,e,w_h)_\text{constr.}$. Similar to ADTW, finding $(p^*,c^*,e^*)$ is not computationally feasible, and instead a suboptimal solution $(p^g,c^g,e^g)$ using hard EM is sought in Algorithm \ref{Algorithm:GARDTW}. $p_v^g$ is obtained by applying RDTW on $s$ and $c^g_{v-1}t+e^g_{v-1}$, and $(c_v^g,e_v^g)$ is computed with equations in Appendix \ref{Appendix:GARDTWEquations}. These equations can be derived by proving convexity of $D_G$ and setting its derivative with respect to $c$ and $e$ to zero. $D_G(s,t,p^g,c^g,e^g,w_h)$ will be referred to as the GARDTW difference measure. From Algorithm \ref{Algorithm:GARDTW} and assuming $n=m$, GARDTW is $O(n_cw_bn)$ in time and $O(w_bn)$ in space, where $n_c$ is the number of iterations to convergence. Similar to ADTW, the scaling $c_v^g$ can be constrained to exist between $c_\text{min}$ and $c_\text{max}$ to avoid improbable scalings.

\begin{algorithm}
\caption{GARDTW}\label{Algorithm:GARDTW}
\begin{algorithmic}[1]
\State $p^g, c^g, e^g, c_0^g \gets 1, e_0^g \gets 0, D_{G,\text{prev}} \gets \infty, v \gets 1$
\While{$1$}
\State $p_{v}^g \gets \underset{p}{\text{argmin}} \ D_G(s,t,p,c_{v-1}^g,e_{v-1}^g,w_h)_\text{constr.}$
\State $(c_v^g,e_v^g) \gets \underset{c,e}{\text{argmin}} \ D_G(s,t,p_v^g,c,e,w_h)_\text{constr.}$
\If {$D_{G,\text{prev}}-D_G(s,t,p^g,c^g,e^g) < D_\text{stop}$}
\State $p^g \gets p_v^g, c^g \gets c_v^g, e^g \gets e_v^g$
\State \textbf{break}
\EndIf
\State $v \gets v + 1$
\EndWhile
\end{algorithmic}
\end{algorithm}

\begin{figure}[!h]
\begin{center}
\includegraphics[scale = 0.38]{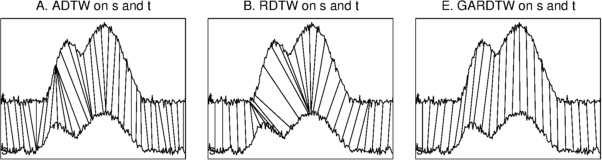}
\caption{ADTW, RDTW and GARDTW on time series scaled differently with component overlap.}
\label{Figure:GARDTWConcept}
\end{center}
\end{figure}

GARDTW is illustrated in Figure \ref{Figure:GARDTWConcept}, and it provides a better alignment than ADTW and RDTW by modeling both scaling and regional emphasis.

\subsection{Local-Affine RDTW}
In local-affine RDTW (LARDTW), the region surrounding each point is assumed to be a scaled and offset version of another region surrounding the corresponding matched point. Formally, LARDTW finds an alignment $p^*$ that minimizes
$$D_L(s,t,p,w_h)=\sum_{k=1}^{|p|}d_l(s_{a_k},t_{b_k},w_h)$$
subject to the DTW constraints, where
\begin{align*}
d_l(s_a,t_b,w_h)=\frac{1}{w_{a,b}}\min_{c_{a,b},e_{a,b}}\sum_{\substack{w=-w_h\\1\leq a+w\leq n\\1\leq b+w\leq m}}^{w_h}d(s_{a+w},t_{b+w}')
\end{align*}
and $t_{b+w}'=c_{a,b}t_{b+w}+e_{a,b}$. $D_L(s,t,p^*,w_h)$ is referred to as the LARDTW difference measure.

The minimizing $(c_{a,b}^*,e_{a,b}^*)$ can be obtained using the following equations for each pair of matched points $(s_a,t_b)$:
\begin{align*}
c_{a,b}^*=\frac{\rho_{a,b}-\frac{1}{w_{a,b}}\phi_{a,b}\tau_{a,b}}{\gamma_{a,b}-\frac{1}{w_{a,b}}\tau_{a,b}^2}, e_{a,b}^*=\frac{1}{w_{a,b}}(\phi_{a,b}-c_{a,b}^*\tau_{a,b})
\end{align*}
where
$$\rho_{a,b}=\sum_{\substack{w=-w_h\\1\leq a+w\leq n\\1\leq b+w\leq m}}^{w_h}s_{a+w}t_{b+w}$$
$$\phi_{a,b}=\sum_{\substack{w=-w_h\\1\leq a+w\leq n\\1\leq b+w\leq m}}^{w_h}s_{a+w},\tau_{a,b}=\sum_{\substack{w=-w_h\\1\leq a+w\leq n\\1\leq b+w\leq m}}^{w_h}t_{b+w}$$
$$\eta_{a,b}=\sum_{\substack{w=-w_h\\1\leq a+w\leq n\\1\leq b+w\leq m}}^{w_h}s_{a+w}^2,\gamma_{a,b}=\sum_{\substack{w=-w_h\\1\leq a+w\leq n\\1\leq b+w\leq m}}^{w_h}t_{b+w}^2$$

\begin{figure}[!h]
\begin{center}
\includegraphics[scale = 0.38]{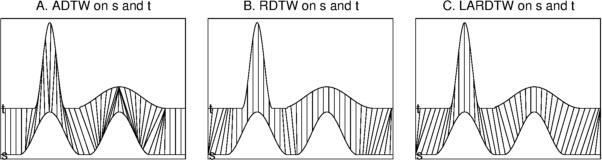}
\caption{ADTW, RDTW and LARDTW on time series with different component scalings and widths.}
\label{Figure:LARDTWConcept}
\end{center}
\end{figure}

Similar to ADTW, the scaling $c_{a,b}$ can be constrained to exist between $c_\text{min}$ and $c_\text{max}$ to avoid improbable scalings. Dynamic programming can again be utilized in the same manner as DTW, where the update formula for constructing the LARDTW table is as follows:
\begin{align*}
& D_L(s,t,p_{(a,b)}^*,w_h) \\ 
&= d_l(s_a,t_b,w_h) + \text{min}(D_L(s,t,p_{(a-1,b-1)}^*,w_h), \\
& \qquad D_L(s,t,p_{(a,b-1)}^*,w_h), D_L(s,t,p_{(a-1,b)}^*,w_h))
\end{align*}
The same backtracking technique used for DTW is applied to the LARDTW table to obtain $p^*$. LARDTW is illustrated in Figure \ref{Figure:LARDTWConcept} where its alignment is visually more appropriate than that of both ADTW and RDTW, and this result is attributed to LARDTW's ability to model different scalings for different regions.

Assuming $n=m$, the LARDTW table requires $O(w_bn)$ elements to be filled with the update formula. Most elements in the LARDTW table can be updated with a time complexity of $O(1)$ instead of $O(w_r)$ using the following observations:
\begin{align*}
\rho_{a,b}=\rho_{a-1,b-1}+s_{a+w_h}t_{b+w_h}-s_{a-w_h-1}t_{b-w_h-1}
\end{align*}
and $(\gamma_{a,b},\tau_{a,b},\phi_{a,b},\eta_{a,b})$ can be updated in a similar manner when $a,b>1$. Thus, $(c_{a,b},e_{a,b})$ can be updated in $O(1)$ time. Furthermore,
\begin{align*}
&d_l(s_a,t_b,w_h)=\frac{1}{w_{a,b}} [ \eta_{a,b} - 2c_{a,b}^*\rho_{a,b} - \\
& \qquad 2e_{a,b}^*\phi_{a,b}+(c_{a,b}^*)^2\gamma_{a,b} + 2c_{a,b}^*e_{a,b}^*\tau_{a,b} + w_{a,b}(e_{a,b}^*)^2 ]
\end{align*}
so $d_l(s_a,t_b,w_h)$ can be updated in $O(1)$ time when $a,b>1$. The time and space complexities for LARDTW are hence analyzed to be $O(w_bn)$.

\section{Evaluation and Discussion}

\subsection{Parameter Values of the Proposed Methods}
The proposed methods of ADTW, RDTW, GARDTW and LARDTW introduce additional parameters to DTW's bandwidth parameter $w_b$. What these parameters are set to or how they are set for evaluation are detailed in Table \ref{Table:ParameterValues}, and these parameters are described below. Recall that the Sakoe-Chiba bandwidth $w_b$ is $1+2w_q$, and $n$ is the typical length of a time series in the evaluated dataset. As a result, $\frac{w_q}{n}$ reflects the bandwidth $w_b$. Similarly, the region width $w_r$ is $1+2w_h$, so $\frac{w_h}{n}$ reflects the region width $w_r$ used by RDTW and methods that augment it. $D_\text{stop}$ is a parameter related to the stopping condition for the ADTW and GARDTW algorithms shown in Algorithm \ref{Algorithm:ADTW} and \ref{Algorithm:GARDTW}. $(c_\text{min},c_\text{max})$ exists to avoid improbable scalings for ADTW, GARDTW and LARDTW. Note that each of these parameters has been defined in the previous sections. Unless mentioned otherwise, for all completed evaluations, parameter values for DTW and our proposed methods were set as reported in Table \ref{Table:ParameterValues}.

\begin{table*}
\caption{Parameters and their assigned values of proposed methods used for evaluation.}
\label{Table:ParameterValues}
\centering
\begin{tabular}{|c|c|c|}
\hline
Parameter & Value & Related methods\\
\hline
$\frac{w_q}{n}$ & Tuned across $\{0,0.05,...,0.5\}$ & DTW, ADTW, RDTW, GARDTW, LARDTW\\
\hline
$\frac{w_h}{n}$ & Tuned across $\{0.05,0.1,...,0.5\}$ & RDTW, GARDTW, LARDTW\\
\hline
$D_\text{stop}$ & $10^{-5}$ & ADTW, GARDTW\\
\hline
$(c_\text{min},c_\text{max})$ & $(0.2,5)$ & ADTW, GARDTW, LARDTW\\
\hline
\end{tabular}
\end{table*}

\subsection{Alignment Evaluation}
The alignments produced by the proposed methods were evaluated by comparing them with true alignments generated through simulations. Two different alignment simulation and evaluation approaches are explored. For global affine simulation, simulated temporal variations, scalings and offsets of real time series were imposed. For component-based simulation, varying widths and scalings were imposed on simulated components, and these components were superimposed with varying temporal offsets to create a set of time series for alignment evaluation.

\subsubsection{Global Affine Simulation}
For global affine simulation, given a real time series $s$, a temporal variant of $s$, $\psi=(\psi_1,...,\psi_z)$, was created by defining a warping function $\omega=(\omega(1),...,\omega(z))$ where $\omega(i)\in \{1,...,n\}$, and setting $\psi_i=s_{\omega(i)}$. The sequence $\omega$ was constrained to be monotonic to maintain a similar structure in $\psi$, and it was modeled as a random sequence in the following manner:
\[
    \omega(i+1)= 
\begin{cases}
	\omega(i)+1&\text{with probability }P_\text{match} \\
    \omega(i)+2&\text{with probability }P_\text{delete} \\
    \omega(i)&\text{with probability }P_\text{insert}
\end{cases}
\]
where $P_\text{match}+P_\text{delete}+P_\text{insert}=1$, $\omega(1)=1$, and this sequence ends when $\omega(z+1)>n$. The true alignment $p^\text{t}=\{(a_1^\text{t},b_1^\text{t}),...,(a_m^\text{t},b_m^\text{t})\}$ between $s$ and $\psi$ can be constructed with $(a_j^\text{t},b_j^\text{t})=(\omega(j),j)$. Interpolation of $\psi$ was completed to have the same length as $s$ and $p^\text{t}$ was modified accordingly for ease of evaluation. Additional scalings and offsets were imposed on $\mu=\bar{c}\psi+\bar{e}$, where $\bar{c}$ and $\bar{e}$ were uniformly distributed with $[\bar{c}_\text{min},\bar{c}_\text{max}]$ and $[\bar{e}_\text{min},\bar{e}_\text{max}]$ respectively. Gaussian white noise with a standard deviation of $\sigma_\text{noise}$ was also imposed. A slight variant of the measure introduced in \cite{keogh2001derivative} was used to evaluate an alignment $p$ on $s$ and $t\sim N(\mu,\sigma_\text{noise}I_{n})$, where $I_n$ is an identity matrix. The variant is as follows:
\begin{align*}
\text{M}_\text{g}(p^\text{t},p)=\frac{1}{\frac{1}{2}n(n-1)}\sum_{i=1}^n\sum_{b_j^\text{t}:a_j^\text{t}=i}\min_{b_l:a_l=i}|b_j^\text{t}-b_l|
\end{align*}
where better alignments have lower $\text{M}_\text{g}$ values.

Global affine simulations were based on 3 datasets taken from \cite{graves2009functional}. These datasets include the number of deaths across different ages in 1 year, the position of the lower lip when saying a certain word, and the temperature across 365 days. All aforementioned datasets are subject to temporal variations, scalings and offsets. For evaluation, 10 time series were taken from each dataset, and 10 time-distorted and affine versions were created for each time series. Let $\sigma$ be the standard deviation of all time series in a specific dataset. Then, $P_\text{match}=0.6$, $P_\text{delete}=P_\text{insert}=0.2$, $\bar{c}_\text{min}=c_\text{min}=0.2$, $\bar{c}_\text{max}=c_\text{max}=5$, $\bar{e}_\text{min}=-\sigma$, $\bar{e}_\text{max}=\sigma$ and $\sigma_\text{noise}=n_l\sigma$, where $n_l$ is defined to be the noise level. An alignment measure $\text{M}_\text{g}$ was obtained for each alignment method and it was averaged within each dataset to get a dataset score. To obtain an appropriate bandwidth and region width, $\frac{w_q}{n}$ and $\frac{w_h}{n}$ were tuned according to Table \ref{Table:ParameterValues} based on the dataset score. The 3 dataset scores were then averaged to obtain $\text{M}_\text{g}^\text{avg}$.

\begin{figure}[!h]
\begin{center}
\includegraphics[scale = 0.45]{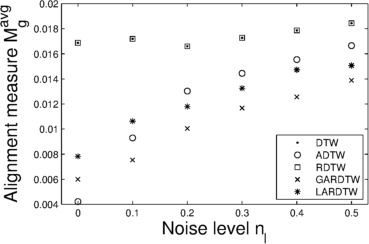}
\caption{Alignment measure $\text{M}_\text{g}^\text{avg}$ of proposed methods across different noise levels.}
\label{Figure:GlobalAlignment}
\end{center}
\end{figure}

$\text{M}_\text{g}^\text{avg}$ values with different methods and noise levels are plotted in Figure \ref{Figure:GlobalAlignment}. When the noise level is low, ADTW outperforms other methods because it accounts for temporal variations, global scalings and global offsets simultaneously without regional emphasis. GARDTW is second behind ADTW when the noise level is low, but it starts to outperform ADTW as the noise level increases because regional emphasis can have an averaging effect in terms of the alignment.

\subsubsection{Component-Based Simulation}
Each time series was comprised of the superposition of different components with varying widths and scalings at different locations for component-based simulation. Let $s$ and $t$ be two simulated component-based time series of length $n$. Also, let $n_c$ be the number of components within a time series. Then, $s$ and $t$ were simulated as follows:
$$s=\sum_{j=1}^{n_c}a_j^{(s)}\phi(i_j^{(s)},w_j^{(s)},z_j),t=\sum_{j=1}^{n_c}a_j^{(t)}\phi(i_j^{(t)},w_j^{(t)},z_j)$$
where $\phi(i,w,z)$ is a component of type $z$ centered at location $i$ with width $w$, and $a$ is the associated scaling factor. Different component types are associated with different windows commonly used in spectral analysis, where $z=1,2,3,4$ denote a Parzen, rectangular, triangular, and flat top weighted window respectively. $z_j$ and $(w_j^{(s)},w_j^{(t)})$ were all generated from discrete uniform distributions with parameters $(z_\text{min},z_\text{max})=(1,4)$ and $(w_\text{min},w_\text{max})=(\frac{n}{2n_c},\frac{n}{n_c})$. $(i_j^{(s)},i_j^{(t)})$ were also generated from a discrete uniform distribution with parameters $(i_\text{min},i_\text{max})=(1,n)$, albeit with an additional constraint that the chronological order of components in $s$ is the same in $t$. In other words, it was assumed that correct alignments can only proceed forward in time as dictated by DTW's monotonicity constraint. In addition, $(a_j^{(s)},a_j^{(t)})$ were generated from a folded normal distribution with parameters $(\mu_a=1,\sigma_a^2)$.

\begin{figure}[!h]
\begin{center}
\includegraphics[scale = 0.45]{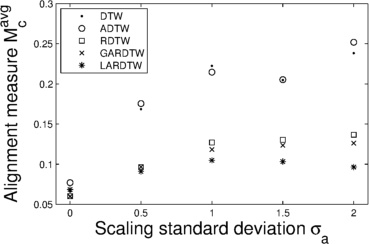}
\caption{Component-based alignment measure $\text{M}_\text{c}^\text{avg}$ of proposed methods across different variances in scaling.}
\label{Figure:ComponentBasedAlignment}
\end{center}
\end{figure}

To produce a true alignment, only points associated with components were considered. The true alignment $p^\text{t}=\{(a_1^\text{t},b_1^\text{t}),...,(a_n^\text{t},b_n^\text{t})\}$ between $s$ and $t$ can be broken down into two parts: non-overlapping and overlapping. For sections that do not have any overlap of components, obtaining the true alignment is straightforward because it is exactly known how to match one component in $s$ to the corresponding component in $t$ during synthesis of the component-based time series. However, confusion arises when component overlap exists, and it was decided to map an overlapped point to the component whose center is closest to the overlapped point, because all simulated components were symmetric and clearly identifiable close to the center. The component-based evaluation measure $\text{M}_\text{c}$ is a slight variation of the measure $\text{M}_\text{g}$ used for global affine simulation:
\begin{equation*}
\text{M}_\text{c}(p^\text{t},p)=\frac{1}{\frac{1}{2}n(n-1)}\sum_{\substack{\forall i \text{ belonging}\\\text{to a component}}}\sum_{b_j^\text{t}:a_j^\text{t}=i}\min_{b_l:a_l=i}|b_j^\text{t}-b_l|
\end{equation*}
The evaluated $\text{M}_\text{c}$ values were averaged across 100 simulated pairs of $s$ and $t$ where $n=400$ and $n_c=4$, and the resulting $\text{M}_\text{c}^\text{avg}$ scores are displayed for each alignment method across different $\sigma_a$ values in Figure \ref{Figure:ComponentBasedAlignment}. For this experiment, $\frac{w_q}{n}=0.5$ and $\frac{w_h}{n}$ were tuned according to Table \ref{Table:ParameterValues} based on $\text{M}_\text{c}^\text{avg}$. It can be observed that methods based on RDTW consistently outperformed DTW and ADTW, because RDTW has a regional emphasis. Furthermore, as the same component is subject to higher variations in amplitude, LARDTW outperforms other methods by larger amounts because each region potentially reflective of a component can be scaled differently in LARDTW. 

\begin{table*}
\caption{Win-loss ratios of 1-NN with proposed difference measures against other state-of-the-art elastic difference measures on the UCR database.}
\label{Table:WinLossDifferenceMeasure}
\centering
\begin{tabular}{|c|c|c|c|c|}
\hline
\multirow{2}{*}{State-of-the-art elastic difference measures} & \multicolumn{4}{c|}{Proposed elastic difference measures}\\
\cline{2-5}
 & ADTW & RDTW & GARDTW & LARDTW \\
\hline
DTW & $1.1$ ($15$,$16$,$12$) & $2.1$ ($25$,$ 8$,$10$) & $2.2$ ($26$,$ 7$,$10$) & $1.1$ ($22$,$ 2$,$19$) \\
\hline
WDTW & $0.7$ ($16$,$ 3$,$24$) & $1.7$ ($25$,$ 4$,$14$) & $2.0$ ($27$,$ 3$,$13$) & $1.0$ ($20$,$ 2$,$21$) \\
\hline
DDTW & $1.9$ ($28$,$ 0$,$15$) & $2.9$ ($32$,$ 0$,$11$) & $2.6$ ($31$,$ 0$,$12$) & $3.3$ ($33$,$ 0$,$10$) \\
\hline
WDDTW & $1.3$ ($24$,$ 1$,$18$) & $2.0$ ($28$,$ 1$,$14$) & $2.4$ ($30$,$ 1$,$12$) & $2.7$ ($31$,$ 1$,$11$) \\
\hline
LCSS & $1.5$ ($25$,$ 1$,$17$) & $3.8$ ($34$,$ 0$,$ 9$) & $3.1$ ($32$,$ 1$,$10$) & $1.3$ ($23$,$ 2$,$18$) \\
\hline
MSM & $1.0$ ($22$,$ 0$,$21$) & $0.8$ ($19$,$ 0$,$24$) & $1.1$ ($23$,$ 0$,$20$) & $0.9$ ($20$,$ 0$,$23$) \\
\hline
TWE & $1.3$ ($23$,$ 2$,$18$) & $1.7$ ($27$,$ 0$,$16$) & $2.0$ ($28$,$ 1$,$14$) & $1.0$ ($21$,$ 1$,$21$) \\
\hline
ERP & $1.4$ ($24$,$ 2$,$17$) & $3.5$ ($33$,$ 1$,$ 9$) & $2.6$ ($30$,$ 2$,$11$) & $1.5$ ($25$,$ 1$,$17$) \\
\hline
\end{tabular}
\caption*{*The number of wins, ties and losses are shown in this order within the brackets against the compared methods.}
\end{table*}

\subsection{Difference Measure Evaluation}
The difference measures associated with the proposed alignment methods were evaluated on 44 datasets from the UCR time series database \cite{keogh2006ucr} using the one-nearest-neighbor (1-NN) error rate. To obtain an appropriate bandwidth and region width for each method (DTW, ADTW, RDTW, GARDTW and LARDTW) and dataset for testing, $\frac{w_q}{n}$ and $\frac{w_h}{n}$ were tuned according to Table \ref{Table:ParameterValues} based on the 2-fold stratified cross-validation error rate on the training set.

\begin{figure}[!h]
\begin{center}
\includegraphics[scale = 0.45]{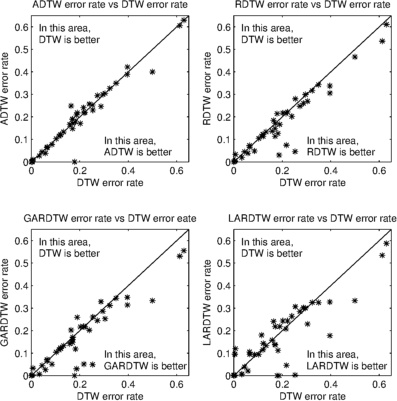}
\caption{1-NN error rates of proposed different measures against DTW on the UCR database. Each point denotes a dataset.}
\label{Figure:ProposedVariantsVSDTW}
\end{center}
\end{figure}

DTW is compared against the proposed difference measures using the 1-NN error rate in Figure \ref{Figure:ProposedVariantsVSDTW}, and clear improvements can be observed for the proposed measures on specific datasets. It is unsurprising that DTW outperformed the proposed measures on certain datasets, because there are datasets (e.g.\ SyntheticControl) where scaling/offset differences and global trends (exact opposites of affine and regional properties) are important for discrimination.

The win-loss ratios of 1-NN with proposed difference measures against other state-of-the-art elastic difference measures are presented in Table \ref{Table:WinLossDifferenceMeasure}, where a tie contributes 0.5 to both the number of wins and the number of losses. We call these compared difference measures elastic because distinct alignments with different properties are also generated in the process of computing these difference measures. It is not unreasonable to expect that the classification performance of the respective difference measures can reflect the quality of their respective alignments. The compared state-of-the-art elastic difference measures include weighted DTW (WDTW) \cite{jeong2011weighted}, derivative DTW (DDTW) \cite{keogh2001derivative}, weighted derivative DTW (WDDTW) \cite{jeong2011weighted}, longest common subsequence (LCSS) \cite{paterson1994longest}, move-split-merge (MSM) \cite{stefan2013move}, time warp edit (TWE) \cite{marteau2009time} and edit distance with real penalty (ERP) \cite{chen2004marriage}. The evaluated results of these compared elastic difference measures on the UCR database were taken from \cite{lines2014time}, and there are 43 datasets that overlap with the evaluation of our proposed methods. Table \ref{Table:WinLossDifferenceMeasure} shows that the RDTW and GARDTW difference measures outperformed DTW with greater than $2$ win-loss ratios. While ADTW and LARDTW do not seem to offer particular advantages over DTW from a win-loss ratio perspective, we will later demonstrate that they offer specialized advantages among the compared difference measures. Furthermore, 1-NN with the proposed difference measures is also competitive with the state-of-the-art elastic difference measures as demonstrated by the associated win-loss ratios.

Among the 43 overlapping datasets evaluated by all compared elastic difference measures, DTW, ADTW, RDTW, GARDTW, LARDTW, WDTW, DDTW, WDDTW, LCSS, MSM, TWE and ERP are each best for 4, 3, 2, 3, 8, 4, 3, 6, 2, 5, 2 and 1 datasets respectively. This suggests that specialized advantages exist for each proposed difference measure on specific datasets even among current state-of-the-art elastic difference measures. To reiterate, among all evaluated measures and datasets in Table \ref{Table:WinLossDifferenceMeasure}, ADTW was found to be best for 3 datasets, RDTW was found to be best for 2 datasets, GARDTW was found to be best for 3 datasets, and LARDTW was found to be best for 8 datasets among the 43 datasets. An ensemble classifier based on the compared elastic difference measures (WDTW, DDTW, WDDTW, LCSS, MSM, TWE and ERP) has been demonstrated to be the most accurate time series classifier ever proposed in the data mining literature \cite{lines2014time}. Considering that our proposed difference measures are best for 16 out of the 43 evaluated datasets among the compared elastic difference measures, it is not unreasonable to expect even better results from incorporating our proposed difference measures into the aforementioned ensemble classifier.

\begin{figure}[!h]
\begin{center}
\includegraphics[scale = 0.48]{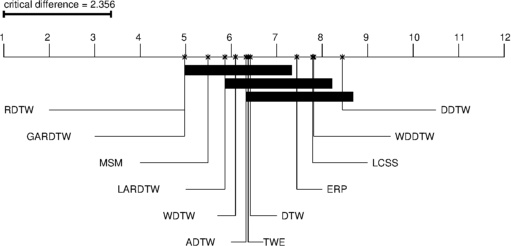}
\caption{Average ranks of 1-NN for proposed difference measures and compared elastic difference measures.}
\label{Figure:ClassifierRank}
\end{center}
\end{figure}

The average performance rankings for 1-NN for the proposed difference measures and the compared elastic difference measures are illustrated in Figure \ref{Figure:ClassifierRank}, where a lower rank corresponds to a more accurate classifier. In this scenario, a total of 12 different elastic difference measures were compared. For each dataset, each compared difference measure was given a rank from $1$ to $12$ based on its associated 1-NN error rate. Finally, for each method, the computed ranks across the 43 UCR datasets were averaged to produce the aforementioned average performance ranking. RDTW and GARDTW clearly have lower average ranks than all other methods. However, it should also be noted that there is no significant statistical difference between any of our proposed methods and most of the compared elastic difference measures based on the Friedman rank test as described in \cite{demvsar2006statistical}. In Figure \ref{Figure:ClassifierRank}, two elastic difference measures are significantly different in rank based on the associated 1-NN error rate if the absolute difference of their average ranks exceeds the critical difference of $2.356$.

\subsection{ADTW vs Normalization Before DTW}
\begin{figure}[!h]
\begin{center}
\includegraphics[scale = 0.45]{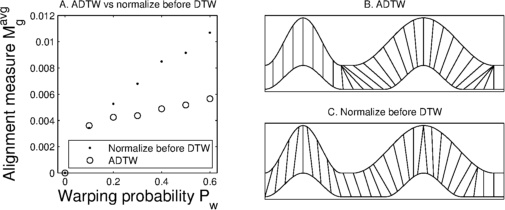}
\caption{ADTW vs normalization before DTW for global affine simulation and illustration of their differences.}
\label{Figure:ADTWvsNormalizeDTW}
\end{center}
\end{figure}

The difference between ADTW and normalization before DTW might not be clear thus far. Normalization before DTW finds the scaling and offset regardless of the alignment, whereas ADTW considers the alignment to find an appropriate scaling and offset in an iterative manner and updates the alignment accordingly. In Figure \ref{Figure:ADTWvsNormalizeDTW}A, ADTW and normalization before DTW are compared across different warping probabilities $P_\text{w}=2P_\text{delete}=2P_\text{insert}$ based on the global affine simulation (described in Section 3.1.1), where ADTW outperforms normalization before DTW by larger amounts as the warping level increases. The difference between their output alignments is also illustrated in Figure \ref{Figure:ADTWvsNormalizeDTW}B and C, and ADTW's alignment is closer to the true one. We are by no means claiming that ADTW is better than normalization before DTW in general, especially since there is no evidence that ADTW-based classification is better than using DTW for classification based on the win-loss ratios for real datasets in Table \ref{Table:WinLossDifferenceMeasure}. However, what we do want to point out is that ADTW can offer distinctly different alignments from normalization before DTW, because ADTW has the unique property where the alignment, scaling and offset are all dependent on each other. This property might be useful when large amounts of warpings occur, as suggested in Figure \ref{Figure:ADTWvsNormalizeDTW}. It is also important to note that ADTW motivated the development of GARDTW and LARDTW.

\subsection{Analysis of Region Width}
\begin{figure}[!h]
\begin{center}
\includegraphics[scale = 0.45]{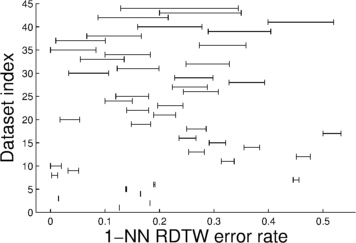}
\caption{Range of 1-NN RDTW error rates across different region widths on each UCR dataset.}
\label{Figure:RDTWRegionWidthSensitivity}
\end{center}
\end{figure}

Figure \ref{Figure:RDTWEffectsOfWidth} demonstrates that region width can be a crucial parameter for RDTW. Figure \ref{Figure:RDTWRegionWidthSensitivity} demonstrates the sensitivity of the 1-NN classification accuracy to the region width for the UCR datasets studied. From Figure \ref{Figure:RDTWRegionWidthSensitivity} it can be seen that the 1-NN RDTW classification accuracy is very sensitive to the region width for some datasets, whereas it is not sensitive to the region width for other datasets. Furthermore, the tuned region width is not arbitrary. In \cite{rakthanmanon2013fast}, FastShapelet provided the lowest error rate ever recorded for the ECGFiveDays dataset by extracting a subsequence (delayed t-wave) confirmed to be discriminative by a medical expert. Interestingly, the tuned region width for RDTW is roughly equal to the length of this subsequence, and LARDTW offers an even better error rate of 0 by appropriately handling the wandering baseline. In \cite{batista2011complexity}, time series extracted from leaf images were presented as a motivating example for a complexity-invariant measure. Associated complexities manifest locally, and the tuned region width for such time series (e.g.\ the OSULeaf dataset) is small, thereby emphasizing these local differences. In this work, 1-NN using LARDTW with a small region width outperforms the complexity-invariant measure for such data. 

\subsection{Actual Runtime Comparison}

In the previous Alignment Methodology section, the runtime complexity of DTW, ADTW, RDTW, GARDTW and LARDTW has been discussed. DTW, RDTW and LARDTW share the time complexity of $O(w_bn)$, whereas ADTW and GARDTW share the time complexity of $O(n_c w_b n)$. Recall that $w_b$ is the Sakoe-Chiba bandwidth, $n_c$ is the number of iterations required for convergence for the ADTW and GARDTW algorithms and $n$ is the length of the time series to find an alignment for.

\begin{figure}[!h]
\begin{center}
\includegraphics[scale = 0.45]{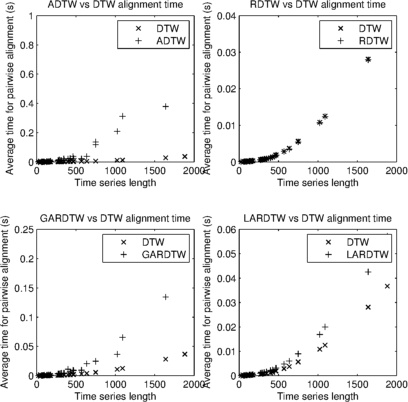}
\caption{Actual time taken to compute the proposed difference measures against DTW on the UCR database. The average pairwise alignment times were calculated by randomly selecting $20$ time series from each UCR dataset for $10$ computations.}
\label{Figure:TimeTaken}
\end{center}
\end{figure}

In Figure \ref{Figure:TimeTaken}, the average time taken to produce an alignment for each proposed method (ADTW, RDTW, GARDTW and LARDTW) is compared against DTW. Each point in Figure \ref{Figure:TimeTaken} denotes a dataset from the UCR database. Note that each dataset from the UCR database has a fixed length. The average pairwise alignment times were calculated by randomly selecting $20$ time series from each UCR dataset for $10$ computations. The parameter values used in this experiment largely follows Table \ref{Table:ParameterValues}, with $\frac{w_q}{n}$ and $\frac{w_h}{n}$ being the exceptions. $\frac{w_q}{n}$ and $\frac{w_h}{n}$ are both set to $0.2$, because most of the $\frac{w_q}{n}$ and $\frac{w_h}{n}$ values tuned on the training sets of the UCR database based on the 1-NN error rate do not exceed $0.2$.

In Figure \ref{Figure:TimeTaken}, we can observe that RDTW has almost identical actual computation time to DTW, but ADTW, GARDTW and LARDTW are visibly slower than DTW. The actual alignment times for LARDTW and DTW does not differ by more than a constant multiple of $0.5$. Both ADTW and GARDTW can be dramatically slower than DTW with more than 5-fold difference for larger datasets because more iterations are required for convergence (reflected in $n_c$). This set of results suggest that ADTW and GARDTW have high relative computation costs when applied to larger datasets, whereas RDTW and LARDTW have computation times comparable to DTW for larger datasets. Nonetheless, ADTW and GARDTW could offer specific advantages for smaller datasets.


%

%
%
%
%

\section{Conclusion}
ADTW, RDTW, GARDTW and LARDTW are alignment methods whose models include affine invariance and regional emphasis. If they are applied to problems whose underlying models are similar to the models behind the methods, the proposed DTW variants are expected to provide performance gains as demonstrated in this work on simulated models and real datasets.

\appendices
\section{GARDTW Affine Equations}
\label{Appendix:GARDTWEquations}
Setting $p=p^g_v$, then
\begin{equation}
\label{Equation:GARDTWEquations}
c^g_v=\frac{\rho-\frac{1}{|p|}\tau\phi}{\gamma-\frac{1}{|p|}\tau^2}, e^g_v=\frac{1}{|p|}(\phi-c^g_v\tau)
\end{equation}
where
$$\rho=\sum_{k=1}^{|p|}\frac{1}{w_{a_k,b_k}}\sum_{\substack{w=-w_h\\1\leq a_k+w\leq n\\1\leq b_k+w\leq m}}^{w_h}s_{a_k+w}t_{b_k+w}$$
$$\gamma=\sum_{k=1}^{|p|}\frac{1}{w_{a_k,b_k}}\sum_{\substack{w=-w_h\\1\leq a_k+w\leq n\\1\leq b_k+w\leq m}}^{w_h}t_{b_k+w}^2$$
$$\tau=\sum_{k=1}^{|p|}\frac{1}{w_{a_k,b_k}}\sum_{\substack{w=-w_h\\1\leq a_k+w\leq n\\1\leq b_k+w\leq m}}^{w_h}t_{b_k+w}$$
$$\phi=\sum_{k=1}^{|p|}\frac{1}{w_{a_k,b_k}}\sum_{\substack{w=-w_h\\1\leq a_k+w\leq n\\1\leq b_k+w\leq m}}^{w_h}s_{a_k+w}$$

\bibliographystyle{ieeetr}
\bibliography{uw-ethesis}

\end{document}